\documentclass[letterpaper, 10 pt, conference]{ieeeconf}  
\pdfoutput=1

\IEEEoverridecommandlockouts                              

\usepackage{graphics} 
\usepackage{epsfig} 
\usepackage{mathptmx} 
\usepackage{times} 
\usepackage{amsmath} 
\usepackage{amssymb}  
\usepackage{subcaption}
\usepackage{gensymb}
\usepackage{xcolor}
\usepackage{url} 
\usepackage{hyperref}
\usepackage{booktabs}
\usepackage{cite}
\usepackage[ruled]{algorithm2e}
\usepackage{comment}
\usepackage{booktabs}
\usepackage{graphicx}
\usepackage{float}
\usepackage{amsmath}
\usepackage{mathrsfs}
\usepackage{mathrsfs}
\usepackage{makecell} 
\usepackage{fancyhdr} 

\title{\LARGE \bf
Neural Trajectory Model: Implicit Neural Trajectory Representation for Trajectories Generation}

\author{Zihan Yu and Yuqing Tang*
\thanks{Zihan Yu is with the Systems Hub, The Hong Kong University of Science and Technology(Guangzhou). This work was conducted while the first author was doing internship at IDEA.
}%
\thanks{Yuqing Tang is with the International Digital Economy Academy.}%
\thanks{*Corresponding author is Yuqing Tang (e-mail: tangyuqing@idea.edu.cn).}%
}
\begin{document}

\maketitle
\thispagestyle{empty}
\pagestyle{fancy}
\fancyfoot[C]{\thepage}  
\renewcommand{\headrulewidth}{0pt}

\begin{abstract}

Trajectory planning is a fundamental problem in robotics. It facilitates a wide range of applications in navigation and motion planning, control, and multi-agent coordination. Trajectory planning is a difficult problem due to its computational complexity and real-world environment complexity with uncertainty, non-linearity, and real-time requirements. The multi-agent trajectory planning problem adds another dimension of difficulty due to inter-agent interaction. Existing solutions are either search-based or optimization-based approaches with simplified assumptions of environment, limited planning speed, and limited scalability in the number of agents. In this work,  we make the first attempt to reformulate single agent and multi-agent trajectory planning problem as query problems over an implicit neural representation of trajectories. We formulate such implicit representation as Neural Trajectory Models (NTM) which can be queried to generate nearly optimal trajectory in complex environments. We conduct experiments in simulation environments and demonstrate that NTM can solve single-agent and multi-agent trajectory planning problems. In the experiments, NTMs achieve (1) sub-millisecond panning time using GPUs, (2) almost avoiding all environment collision, (3) almost avoiding all inter-agent collision, and (4) generating almost shortest paths. We also demonstrate that the same NTM framework can also be used for trajectories correction and multi-trajectory conflict resolution refining low quality and conflicting multi-agent trajectories into nearly optimal solutions efficiently. (Open source code is available at \url{https://github.com/laser2099/neural-trajectory-model})
\end{abstract}

\section{INTRODUCTION}
Trajectory planning plays a crucial role in the application of autonomous agents. It requires the accurate representation of the environment and the generation of desired trajectory solutions by environmental and performance constraints. Commonly, researchers employ discretization maps, such as grid maps or graphs, to represent the planning environment. These methods simplify complex environments; however, discretization also leads to a loss of information when representing finer details in the environment and can become inflexible as the size and complexity of the environment increase. Confronting these limitations, our work adopts a differentiable environment representation approach, especially a differential signed distance field model to represent the planning environment continuously, offering a more flexible and accurate representation of complex environments.

Trajectory planning research is usually conducted from two perspectives -- single-agent and multi-agent trajectory planning. Single-agent trajectory planning entails finding a trajectory between two positions in a space and is a fundamental, well-researched issue in artificial intelligence. Optimal solutions to trajectory planning problems are typically achieved using search algorithms such as the A* series algorithms~\cite{hart1968formal} and the RRT series algorithms~\cite{aoude2010threat,yu2021npq}.  Multi-agent trajectory planning (MATP) is an extension of single-agent trajectory planning that involves multiple agents. This problem comprises an environment and several agents, each with a unique starting and ending state. The objective is to determine paths for all agents from their respective starting points to their goals, ensuring that agents do not collide with the environment and other agents during their movements~\cite{hou2022enhanced}. To solve the MATP problem, many state-of-the-art require the use of searching-based algorithms to generate reference paths for further generation of multi-agent trajectories. For example, EDG-Team~\cite{hou2022enhanced} implemented Enhanced-Conflict-based search (ECBS)~\cite{barer2014suboptimal} algorithm as the front-end of group planning. A*+ID~\cite{standley2010finding} performs A* searches in priority order to solve the combinatorial explosion problem.

Although search-based algorithms exhibit satisfactory performance in certain scenarios, they can be highly time-consuming when applied to single or multi-agent trajectory planning problems. For each agent and every instance, the trajectory planning problem is treated as a completely new challenge, requiring repeated searches for potential solutions within the environment. However, many trajectory planning problems share striking similarities. As their proximity in starting and ending positions as well as their identical environments. Thus, effectively leveraging previous planning experience could lead to more efficient problem-solving.

In this paper, we make the first attempt to reformulate single-agent and multi-agent trajectory planning problems as query problems over an implicit neural representation of trajectories.  We leverage the latest advancement in the implicit neural representation of complex structures to efficiently represent valid and nearly optimal trajectories in a given environment. Then we cast the trajectory planning, refinement, and coordination problems as query problems over the implicitly represented trajectory models. The neural trajectory model can implicitly encode seen trajectories and generalize to unseen trajectories while preserving the validity constraints and optimality qualities. We conduct experiments in three distinct environments and demonstrate that our approach is at least 10 times faster than baseline methods while maintaining the desired planning performance. The contributions of our work can be summarized as follows:
\begin{itemize}
\item We present a novel reformulation of single-agent and multi-agent trajectory planning problems as query problems with implicit neural representation of valid and optimal trajectories in a given environment.
\item Our method demonstrates fast planning speed and superior performance for both single-agent and multi-agent trajectory planning.
\item We also demonstrate that the same neural formulation can be employed for multi-agent coordination and trajectories de-conflict. 
\item Preliminary experiments also demonstrate the potential of this approach towards solving large-scale multi-agent trajectory planning problems.
\end{itemize}

\section{RELATED WORK}
\subsection{Search-based single agent trajectory planning}
Search-based path planning algorithms are widely used in single-agent trajectory planning problems due to their efficiency and ease of implementation. A pioneering work in the field of path planning was proposed by E. W. Dijkstra~\cite{dijkstra2022note} in 1959, which demonstrated how to find the shortest path between two nodes on a graph. Subsequently, algorithms such as A*, and RRT were proposed to solve various problems. Over the past several decades, numerous improved versions have been presented, including Hybrid A*~\cite{dolgov2008practical}, RRT*~\cite{karaman2011sampling}, and LPA*~\cite{koenig2002incremental}. Although these improved algorithms exhibit better performance compared to their original counterparts, running time remains a significant concern when implementing search-based algorithms in trajectory planning, as they require repeated searches in the space. Therefore, in our work, we employ a neural trajectory model that capitalizes on the latest advancements in neural models and trajectory planning experiences to address the problems in a novel manner circumventing redundant searches and attaining significantly enhanced performance.

\subsection{Multi-agent trajectory planning}
The multi-agent trajectory planning (MATP) problem is an extension of the single-agent trajectory planning problem. Numerous researchers attempt to solve such problems by leveraging single-agent search-based algorithms. When no conflicts arise, each agent employs single-agent planning algorithms independently; otherwise, a new policy is implemented to coordinate conflicts among the agents. \cite{standley2010finding} conducts A*-based searches and then applies independence detection to resolve potential collisions. M*~\cite{wagner2011m} dynamically adjusts the search dimension based on the presence or absence of conflicts. In conflict-free situations, each agent takes the best possible action on its own; otherwise, conflict-based methods transform conflicts into constraints added to agents to tackle the problem. The Conflict-based search (CBS) algorithm~\cite{sharon2015conflict} conducts searches on a constraint tree (CT). When a conflict arises, the high level creates a new node and converts the conflict into a constraint added to the agents, while the low-level agents perform A* searches. Once all agents reach their respective targets without conflicts, the node becomes the final solution. EDG-TEAM~\cite{hou2022enhanced} employs ECBS to generate initial paths and then implements a group-based optimization to further ensure the algorithm's performance.

While such methods address certain issues, they involve simplifying assumptions about the environment, limited planning speed, and a restricted number of agents in multi-agent planning settings. In this work, we avoid extensive searching and employ a neural trajectory model instead to tackle these problems.

\subsection{Neural Models}
Implicit neural environment representations employ neural networks to represent the geometry, and occasionally the color and texture, of intricate 3D scenes. Typically, these representations utilize a labeled dataset to learn a function in the form of $f_\theta(p) = \sigma$, where $f$ denotes a neural network parameterized by the weights $\theta$, $p$ signifies a low-dimensional query such as a  $(x, y, z)$ 3D coordinate, and $\sigma$ represents a relevant quantity, usually scalar. Existing neural implicit representations' working mechanisms can be classified into generalizable~\cite{park2019deepsdf, chabra2020deep} and overfit paradigms~\cite{davies2020effectiveness, zhang2023neurogf, sitzmann2020implicit, takikawa2021neural}. The latter approach focuses on accurately reproducing a single shape by deliberately overfitting a single neural network model. In this work, we adopt the implicit representation paradigm of neural models. Specifically, we extend the neural fields formulation to accommodate trajectories --- overfitting a multi-trajectory function for a given environment so that valid and nearly optimal trajectories of the environment can be queried with their start-end positions as input.


\section{Neural Trajectories}
\label{sec:neural-traj}
\subsection{Problem Formulation}
In this section, we present the problem formulation for generating safe trajectories in a multi-agent robotic system comprising $N$ agents. The task assigned to the $i$th agent is to move from a starting position $s^i$ to a goal position $g^i$. The environment is represented by a Signed Distance Field (SDF)~\cite{malladi1995shape}. $SDF(x)$ determines the distance to the closest surface of any 3D coordinate $x$ in the environment where (1) $SDF(x) < 0$ means $x$ is inside an object; (2) $SDF(x) = 0 $ means $x$ is on the surface of an object; (3) $SDF(x) > 0$ means $x$ is outside any objects in free space. Such an SDF can be derived from a 3D mesh model, occupancy grids, CSG (constructive solid graph) and other 3D environment representations. 

A trajectory $\mathbb{W}^i$ for agent $i$ is denoted as a set of time-stamped waypoints:
$$
    \mathbb{W}^i=\{(t_i^0, p_i^0),(t_i^1,p_i^1),(t_i^2,p_i^2),...,(t_i^{T-1},p_i^{T-1}),(t_i^T,p_i^{T})\}
$$ 
where (1) the 4D waypoint $w_i^j = (t_i^{j}, p_i^{j}$) is the $j$th waypoint ($j \in \{0, ..., T\}$) with $t_i^{j}$ and $p_i^{j}$ being the $j$th timestamp and $j$th coordinate in trajectory; (2) $s_i=p_i^0$ and $g_i=p_i^T$ are the starting and the goal positions; (3) $T$ is the time horizon for the trajectory. 

A trajectory $\mathbb{W}^i$ is considered environmental collision-free if all its waypoints are in free space in the environment with a safety threshold $S_{thresh}$: i.e. $SDF(p_i^j) > S_{thresh}$ for all $p_i^j \in \mathbb{W}^i$. Two trajectories $\mathbb{W}^i$ and $\mathbb{W}^k$ are considered inter-collision free if any pair of waypoints $(t_j^i, p_j^i) \in \mathbb{W}^i$ and $(t_{j'}^k, p_{j'}^k) \in \mathbb{W}^j$ are conflict-free: (1) time separated with a threshold $t_{thresh}$: $\lvert t_j^i - t_{j'}^k \rvert > t_{thresh}$; or (2) distance separated (Euclidean distance) with a safety threshold $dis_{thresh}$: $\lVert p_j^i - p_{j'}^k\rVert > dis_{thresh}$.

A multi-agent trajectory generation problem is to generate a set of trajectories $\mathbb{W} =\{\mathbb{W}^1, \mathbb{W}^2,..., \mathbb{W}^N\}$ given the starting and goal positions $\{(s^1,g^1), (s^2,g^2), ..., (s^N,g^N)\}$ of $N$ agents. The requirement is that all trajectories in $\mathbb{W}$ are both environmental collision-free and inter-collision-free. We seek to learn a neural function $f_{\Theta}$ parameterized by $\Theta$ to generate such a trajectory set:
\begin{equation}
      \mathbb{W} = f_{\Theta}(\{(s^1,g^1),(s^2,g^2),...,(s^N,g^N)\}).
       \label{eq:neural-traj-model}
\end{equation}

\subsection{Neural Trajectory Model}
%
\begin{figure*}[htbp]
  \centering
  \includegraphics[width=0.85\textwidth]{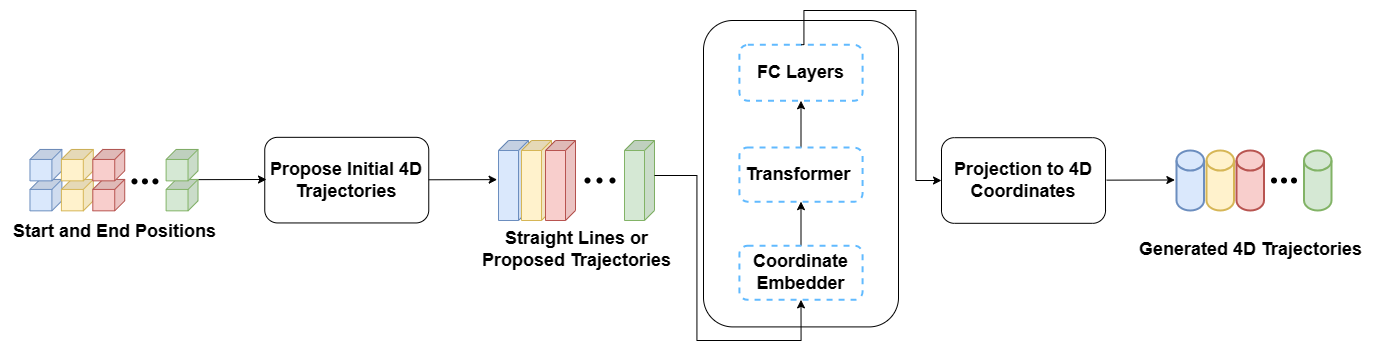}
  \caption{Neural Trajectory Model}
  \label{fig:neural-trajectory-model}
\end{figure*}
Our neural trajectory model is designed for both single-agent and multi-agent trajectory generation problems.  As shown in Figure~\ref{fig:neural-trajectory-model}, the model accepts a set of $N$ start-end positions as input and output $N$ corresponding 4D trajectories. A 4D trajectories proposing step is first employed to propose an initial set of 4D trajectories for $N$ start-end positions. We implement a simple trajectory proposal method using line interpolation. $T$ discrete 3D points are sampled with uniform intervals along a straight line between the start and end position $(s, g)$ as follows: $l_{s\to g}=\left\{ p_j = s+(g-s)/T\times j \right\}_{j=0}^T$ where $l_{s\to g} \in \mathbb{R}^{(T+1) \times 3}$ serves as the initial path proposal. Equally separated time-stamps $\{t_0, ..., t_T\}$ are sampled between $0$ and $T$ and prepend to the path proposal  $l_{s\to g}$ to form the trajectory proposal: 
\begin{equation}
 L_{s \to g} = \{ (t_j, p_j) \}_{j=0}^T
        \label{eq:init-proposal}
\end{equation}
%
%
Other initial trajectories proposal methods can also be used. For example, in Section~\ref{sec:coordinate-experiments} we demonstrate to feed trajectories created by other methods or a single-agent neural trajectory model into the transformer as initial trajectories to generate desired high-quality trajectories. The 4D coordinates in the proposed trajectories are then fed to a coordinate embedder $CoordinateEmbedder$ to obtain their corresponding high dimensional embeddings. We adapt the positional embedding approach in the original transformer paper \cite{vaswani2017attention} to 4D positional embeddings. 
\begin{equation}
       E_l=\textit{CoordinateEmbedder}(L_{s\to g}),
       \label{eq:corod-embed}
   \end{equation}
Coordinate embeddings are then fed into a transformer~\cite{vaswani2017attention}, conducting cross attention mechanism to capture the relationship among the trajectories. At the end, a fully-connection layer is applied to project the transformer's  high dimensional vector sequence outputs to 4D coordinates sequences to form the $N$ desired 4D trajectories of time horizon $T$ for $N$ agents:
 %
\begin{equation}
       f_{\Theta} =\textit{FC}(\textit{transformer}(E_l)).
       \label{seq:neural-traj-model}
\end{equation}
The focus of this work is on creating neural trajectory models for a static environment. This framework can be easily extended to accommodate a dynamic environment by adding perception embedding context to the input of the transformer. We will leave this for our future work.

\subsection{Model Training}
\label{sec:model-training}
%
\begin{figure*}[htb]
  \centering
  \includegraphics[width=0.85\textwidth]{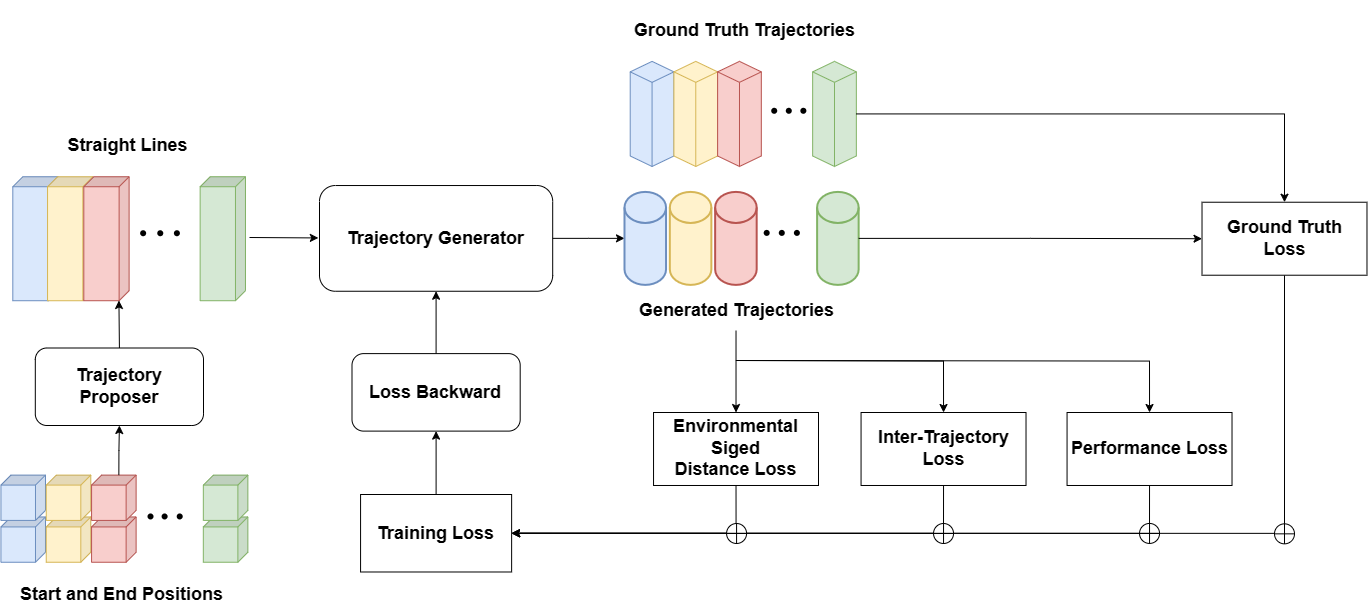}
  \caption{Neural Trajectory Model Training}
  \label{fig:neural-trajectory-training}
\end{figure*}
As shown in Figure~\ref{fig:neural-trajectory-training}, the neural trajectory model is trained with ground truth trajectories data along with additional regularization of collision-free and performance metrics (.e.g. travel distance, time, etc.) requirements. The training data set $D$ is of the from $D = \{ ( SG_j,  \mathbb{W}_j) \}$ where each $SG_j$ is $N$ start-goal positions and each $\mathbb{W}_j$ is the corresponding ground-truth multi-agent trajectories which satisfy collision-free and the desired performance requirement. The start and goal positions of ground truth trajectories are sampled from a given environment. The goal of the training process is to enable the model to generalize the trajectory generation capability beyond the sampled start-goal positions to unseen start-goal positions of $N$ agents in the same environment. We will introduce the training loss functions first and describe the ground truth data generation process in Section~\ref{sec:data-generation}.

\subsubsection*{\bf Ground Truth Supervision}
We compute the point-wise L1 losses between the model generated trajectories and ground-truth trajectories to enable fast convergence to plausible trajectories for the environment.
\begin{equation}
       l_{gt}=\frac{1}{T \times N} \sum_{i=1}^{N}\sum_{j=1}^{T}(\Vert \hat{w}^i_j - w_j^i \Vert_1)
       \label{eq:ground-truth-loss}
   \end{equation}
where $\hat{w}^i_j$ is the $j$th model generated trajectory waypoint and $w_j^i$ is the corresponding ground truth for the $i$th agent. 

\subsubsection*{\bf Environmental Safety Loss} We point-wisely compute signed distance values of trajectories waypoints to ensure them to be a distance away from environment obstacles for a safety distance threshold. We adapt the differential SDF implementation in the Kaolin library~\cite{tsang2022kaolinLibrary} to enable loss back-propagation of the environmental safety loss. Neural SDFs \cite{park2019deepsdf} can also be used here.
\begin{equation}
       l_{sdist}=\frac{1}{T \times N} \sum_{i=1}^{N}\sum_{j=1}^{T} max\{(S_{thresh}-SDF(p^i_j), 0\} .
       \label{eq:env-loss}
   \end{equation}

\subsubsection*{\bf Inter-Trajectory Conflict Loss}
We ensure distances between any two waypoints of two trajectories are separated by at least $w_{thresh} = \langle t_{thresh}, dist_{thresh} \rangle$ by the following inter-collision loss:
\begin{equation}
       l_{inter}=\frac{1}{T \times N} \sum_{i,k=1}^{N}\sum_{j,j'=1}^T max(0, w_{thresh} - \Vert w_j^i - w_{j'}^k).
       \label{eq:inter-loss}
\end{equation}

\subsubsection*{\bf Performance Loss} We ensure the generated trajectories to be better than the ground-truth length $\hat{d}^i$ as an example performance metric loss: 
\begin{equation}
       l_{dist}=\frac{1}{N}\sum_{i=1}^{N} max(0, \hat{d}^i - d^i )
       \label{eq:distance-loss}
   \end{equation}
where $\hat{d}^i$ and $d^i$ are the ground truth and the predicted travel distance of agent $i$ respectively.

The overall training objective is as follows:
\begin{equation}
       loss=\lambda_1l_{gt} + \lambda_2l_{sdist} + \lambda_3 l_{inter} + \lambda_4 l_{dist}
       \label{eq:neural-traj-loss}
   \end{equation}
where $\lambda_1$, $\lambda_2$, $\lambda_3$ and $\lambda_4$ are the corresponding combination weights. 


\subsection{Neural Trajectory Inference}
\label{sec:model-inference}
%
The inference process of NTG is shown in Algorithm~\ref{alg:neural-traj-inference}. The neural trajectory inference takes start-goal positions as input and generates desired trajectories for $N$ agents. It consists of two main steps and one optional optimization step. The first two steps are trajectory propose generation and neural model processing. An optional optimization can be employed to further optimize the output. It can be either another round of neural trajectory model application or using the trajectory optimizer described in section~\ref{sec:traj-optimizer}.

\begin{algorithm}
		\caption{Neural Trajectory Inference}
        \label{alg:neural-traj-inference}  
		\LinesNumbered 
		\KwIn{Start-Goal positions: $SG_{pairs}=\{(s^1,g^1),(s^2,g^2),...,(s^N,g^N)\}$ }
		\KwOut{Trajectories: $\mathbb{W}=\{\mathbb{W}^1, \mathbb{W}^2,...,\mathbb{W}^N \}$}
        $InitTrajs \leftarrow  \textit GetLines(SG_{pairs})$ \;
        $\mathbb{W} \leftarrow \textit NeuralTrajectory(InitTrajs)$ \;
        \If{$further\_optimize$ \textbf{is} $True$}{
            \tcc{Optional trajectory optimization step}
            $\mathbb{W} \leftarrow \textit TrajectoriesOptimizer(\mathbb{W})$ \;
        }
        \Return $\mathbb{W}$\;
       
\end{algorithm}

\subsubsection*{Multi-trajectories Coordination and De-Conflicting}
The neural trajectory model can not only support trajectory planning but also trajectory de-conflicting. Instead of a straight line between the start and end positions as input, in the coordination task, the inference procedure in Algorithm~\ref{alg:neural-traj-inference} can take generated trajectories with conflicts or non-optimal trajectories as input and output almost conflict-free and optimal trajectories.



\subsection{A Trajectory Optimizer}
\label{sec:traj-optimizer}
\begin{algorithm}
		\caption{Trajectories Optimizer }
        \label{alg:traj-optimizer}
		\LinesNumbered 
		\KwIn{Initial Trajectories: $\mathbb{W}^0 =\{\tau^1, \tau^2,...,\tau^N\}$  }
		\KwOut{Optimized Trajectories: $\mathbb{W}^*$}
         $\mathbb{W} \gets \mathbb{W}^0 $ \;
        \Repeat{desired trajectories are obtained or time-out}{
             \tcc{$w_j^i$s are 4D waypoints in $\mathbb{W}$}
            $\nabla_{w_i^j} \gets \frac{\partial }{\partial w_i^j } J(\mathbb{W}) $ \label{ln:trajs-gradient}\; 
            $\mathbb{W} \gets \mathbb{W} - \gamma \nabla_{w_i^j} $ \label{ln:update-step}\;
        } 
        \Return $\mathbb{W}$\;
\end{algorithm}

The Trajectory Optimizer as specified in Algorithm~\ref{alg:traj-optimizer}  makes use of the following loss functions to optimize proposal trajectories: 
\begin{equation}
       J=  \lambda_1 l_{sdist} + \lambda_2 l_{inter} + \lambda_3 l_{dist}
       \label{s_equation}
\end{equation}
where $\lambda_1$, $\lambda_2$ and $\lambda_3$ are the corresponding combination weights. $l_{sdist}$ (Equation~\ref{eq:env-loss}), $l_{inter}$ (Equation~\ref{eq:inter-loss}), and $l_{dist}$ (Equation~\ref{eq:distance-loss} ) are the environmental loss, the inter-trajectory collision loss and the performance loss defined in Section~\ref{sec:model-training}. In  Algorithm~\ref{alg:traj-optimizer} line~\ref{ln:trajs-gradient} takes gradients with respect to trajectories waypoint coordinates instead of taking gradients with respect to neural model parameters as in Section~\ref{sec:model-training}. $\gamma$ in Line~\ref{ln:update-step} is the gradient descent update learning rate. The trajectories refined by the trajectory optimizer are nearly collision-free with desired performance, such as shorter travel distances.

\subsection{Training Data Generation}
\label{sec:data-generation}
To support the model training process described in Section~\ref{sec:model-training}, we devise the data generation process as in Algorithm~\ref{alg:training-data-geration}. It requires an environment model $ENV$ from which we can sample 3D coordinates that are outside any obstacles in the environment in the first place. We sample from the environment $M$  start-goal positions. Then we apply a trajectory proposer (Line~\ref{ln:traj-prop}) to generate initial trajectories for optimization. We use multi-agent A-Star algorithms \cite{hart1968formal} as our trajectories proposer--- in particular, we use the implementation available in \cite{adamkiewicz2022vision}. Then the trajectory optimizer described in Algorithm~\ref{alg:traj-optimizer} is applied to optimize the trajectories to obtain nearly optimal trajectories with no environmental and inter-agent collisions.

\begin{algorithm}
    \caption{Training Data Generation Process}
    \label{alg:training-data-geration}
    \LinesNumbered 
    \KwIn{An environment SDF model $SDF$}
    \KwOut{Trajectories Dataset: $D = \{ ( SG_j,  \mathbb{W}_j) \}$}    
    \Repeat{$M$ data instances are obtained or time-out}{
        Sample $SG_{j}=\{(s^1_j, g^1_j),...,(s^N_j,g^N_j) \mid SDF(s_j^i) > 0, SDF(g_j^i) > 0 \}$  from the environment model $ENV$  \;
        $\mathbb{W}^0 \gets TrajectorProposer(SG_j)$ \label{ln:traj-prop} \;
        $\mathbb{W}_j \gets TrajectoryOptimizer(\mathbb{W}^0)$ \;
        $D \leftarrow D \cup \textit \{(SG_j, \mathbb{W}_j)\}$ \;
    }
    \Return $D$
\end{algorithm}


\section{Experiments}
\label{sec:exp}
In this section, we conduct detailed evaluation tests on the proposed methods. All tests are run on a device equipped with an Intel(R) Xeon(R) Gold 5218 CPU and an RTX 3090 GPU. 

\subsection*{Experiment Environments}
We use 3D mesh to represent the environment. Detailed mesh statistics can be found in Table ~\ref{tab:3d-envs}.
\begin{table}[ht]
\renewcommand\arraystretch{2}
\caption{Statistics of 3D Experimental Environments}
    \centering
    \begin{tabular}{c|c c}
    \hline
         Environment &\makecell[c]{Number of \\Vertices} & \makecell[c]{Number of \\Obstacles} \\
         \hline
         Stonehenge& 10995 &24 \\
         \hline
         Ice Forest& 10692 & 25 \\
         \hline
         Building Forest& 20846 & 60 \\
         \hline
    \end{tabular}
    \label{tab:3d-envs}
\end{table}
%
\subsection*{Evaluation Metrics}
We use the following evaluation metrics to quantitatively compare our approaches with baselines.

\noindent\textbf{Environmental Collision Rate (ECR)}: A collision is defined as the agent being closer to an obstacle than the safety distance. If a trajectory contains a collision, it is considered a collision trajectory. ECR is the ratio of the number of collision trajectories to the total number of trajectories. 

\noindent\textbf{Inter Collision Rate (ICR)}: An inter-trajectory collision occurs when two agents have similar timestamps (usually within a threshold, such as 0.1s) and their distance is less than the safety distance. ICR is the ratio of the number of inter-trajectory collisions to the total number of trajectories.

\noindent\textbf{Travel Distance (TD)}: Travel Distance refers to the average length of all trajectories in a trajectory planning task. 

\noindent\textbf{Calculation Time (CT)}: Calculation time, measured in seconds (s), is the time required to generate trajectories for agents.

It is also important to note that although the search-based baselines are sound, they don't guarantee completeness for collision-free paths. They provide valid paths when reaching solutions of no collisions, but fail to offer solutions after exceeding pre-set computational cost budget. In our study, when computing collision rates, we treated unsolvable cases within given computational cost as collision cases.

\subsection*{Data}
We employ the data generation process described in Section~\ref{sec:data-generation} to generate ground truth data for our experiments. Their statistics are shown in Table~\ref{tab:ntg-datasets}.

\begin{table}[h]
\renewcommand\arraystretch{2}
    \caption{Experiment Datasets}
    \centering
    \begin{tabular}{c|c c}
      \hline
       \textbf{Environment}  &  \textbf{Single-agent} & \textbf{Multi-agent} \\
       \hline
       Stonehenge training data &  512  & 324 \\
        \hline
       Stonehenge validation data &  110  & 51 \\
        \hline
       Stonehenge test data   &128  & 72 \\
     \hline
       Ice-forest training data   & /  & 1003 \\
        \hline
       Ice-forest validation data  & /  & 197 \\
        \hline
       Ice-forest test data  & /  & 203 \\
     \hline   
       Building-forest training data  &  /  & 1710 \\
        \hline
       Building-forest validation data  & /  & 317 \\
        \hline
       Building-forest test data  &  /  & 330 \\
     \hline       
     \end{tabular}

    \label{tab:ntg-datasets}
\end{table}
\subsection{Single Agent Trajectory Planning}
The neural trajectories can be applied to perform single-agent trajectory planning. We conduct single-agent experiments in the Stonehenge environment from \cite{adamkiewicz2022vision}. We compare the proposed method with Nerf-Nav~\cite{adamkiewicz2022vision}, an advanced planning algorithm and the classical search-based algorithm A*~\cite{hart1968formal} (implementation of \cite{adamkiewicz2022vision}). In this scenario, the NTM is trained with $512$ trajectories. We evaluate the models on 128 start and end position pairs, which were randomly sampled from the environment, excluding the training data. The evaluation results are presented in Table ~\ref{tab:single_agent}.
\begin{table}[h]
\renewcommand\arraystretch{2}
		\begin{center}
		\caption{Single-agent Experiments in Stonehenge Env}
			\begin{tabular}{c| c c c}
				\hline
				\textbf{Method} & \textbf{\makecell[c]{Environmental \\Collision Rate}}&\textbf{\makecell[c]{Travel \\Distance}}& \textbf{\makecell[c]{Calculation \\Time(s)}} \\
				\hline
				Nerf-Nav~\cite{adamkiewicz2022vision} &0.0653&1.2567&2.0\\
				\hline
     A*~\cite{hart1968formal} &0.104 &\textbf{1.098} &0.1840 \\
				\hline
			\textbf{NTM(Ours)}&\textbf{0.054}&1.1448&\textbf{0.0025}\\
				\hline
    Ground Truth &0.0 &1.1011 &/ \\
    \hline
			\end{tabular}
			\label{tab:single_agent}
		\end{center}
	\end{table}

From the results, we can see that NTM can generate trajectories much faster, resulting in shorter and safer routes compared to the baseline methods.

\subsection{Multi-agent Trajectory Planning}
For the MATP problem, we experiment with an eight-agent trajectory planning problem for evaluation. We compare our method with a SoTA approach EDG-TEAM~\cite{hou2022enhanced} and a classical MATP algorithm ECBS~\cite{barer2014suboptimal}. Evaluations are conducted across three distinct environments: Stonehenge~\cite{adamkiewicz2022vision}, Ice Forest~\cite{hou2022enhanced}, and Building Forest Environment (re-created and adapted from \cite{Free3D}).


\begin{figure}[ht]
  \centering
  \includegraphics[width=0.8\linewidth]{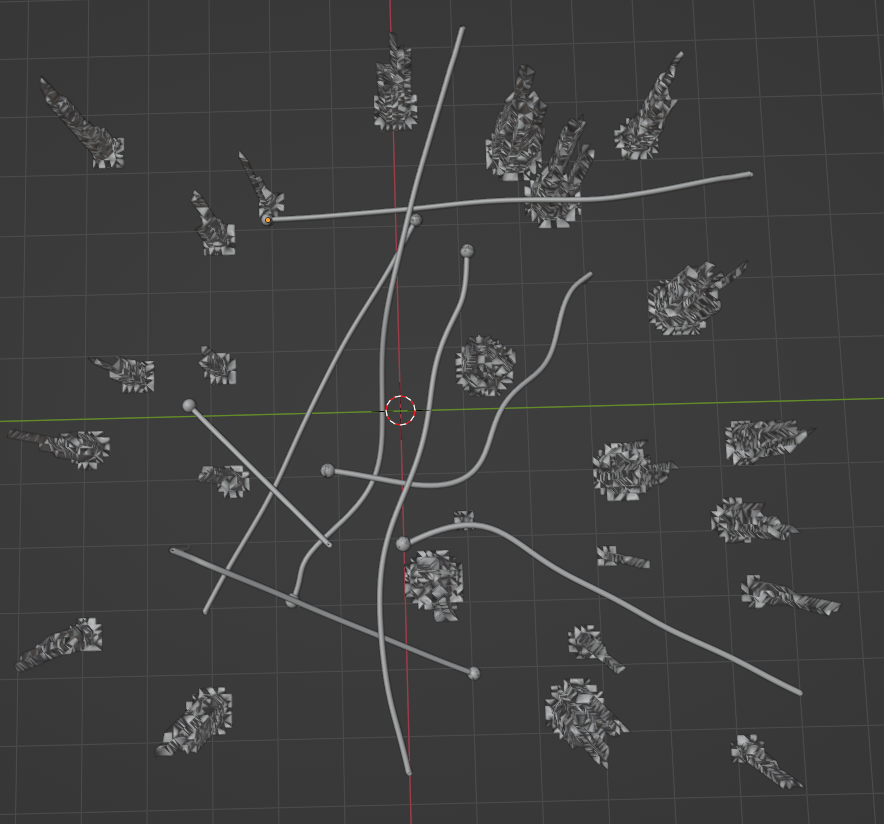}
  \caption{Eight-agent trajectory planning in Ice Forest Env}
  \label{fig:ice-forest-8agents}

\end{figure}

\begin{figure}[htbp]
  \centering
  \includegraphics[width=0.8\linewidth]{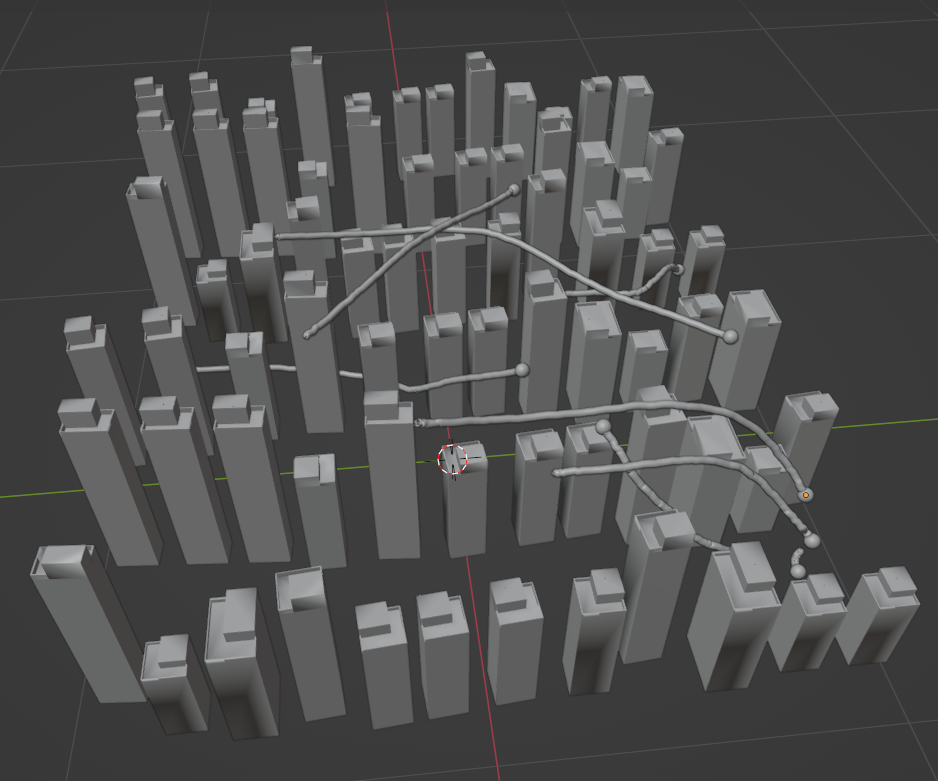}
  \caption{Eight-agent trajectory planning in Building Forest Env}
  \label{fig:building-forest-8agents}

\end{figure}

For the Stonehenge environment, we train the NTM on 324 trajectories and test it on 72 distinct start-end position pairs. In the ice forest environment, the NTM is trained on 1003 trajectories and evaluated on 245 unique start and end position pairs. Lastly, in the building forest environment, the NTM is trained on 1710 trajectories and assessed on 330 different start and end position pairs. The testing start and end position pairs are randomly chosen from the environment, excluding the training data. The results are provided in Table \ref{tab:results}.

\begin{table*}[t]
    \centering
    \caption{Comparison of eight-agent planning in a Stonehenge, Ice Forest, Building Forest environment}
\scalebox{1.1}{
    \begin{tabular}{@{}lccccc@{}}
    \toprule 
    & \multicolumn{4}{c}{\textsc{Experiment in Stonehenge environment}} \\
    \cmidrule(lr){2-5} 
    & \textbf{Inter Collision Rate}&\textbf{\makecell[c]{Environmental \\Collision Rate}}& \textbf{Travel Distance}   &\textbf{Calculation Time(s)}\\
    \midrule
        EDG~\cite{hou2022enhanced}& 0.048 & 0.312 & \textbf{1.705}  & 0.017 \\ 
            ECBS~\cite{barer2014suboptimal}& 0.088 & 0.325 & 1.992  & 0.21 \\ 
           \textbf{NTM (Ours)} & \textbf{0.032} & \textbf{0.027} & 1.877  & \textbf{0.0025} \\ 
            \midrule
            Ground Truth & 0.0 & 0.0 &  1.598  & / \\ 
            \bottomrule
    \toprule 
    & \multicolumn{4}{c}{\textsc{Experiment in Ice forest environment}} \\
    \cmidrule(lr){2-5} 
    & \textbf{Inter Collision Rate}&\textbf{\makecell[c]{Environmental \\Collision Rate}}& \textbf{Travel Distance}   &\textbf{Calculation Time(s)}\\
    \midrule
        EDG~\cite{hou2022enhanced}& 0.033 & 0.146 & 5.717  & 0.022 \\ 
            ECBS~\cite{barer2014suboptimal}& 0.0412 & 0.187 & 5.871  & 0.32 \\ 
            \textbf{NTM (Ours)} & \textbf{0.022} & \textbf{0.0} & \textbf{5.179}   & \textbf{0.0021} \\ 
            \midrule
            
            Ground Truth & 0.004 & 0.004& 5.109& /   \\ 
            \bottomrule
   \toprule 
    & \multicolumn{4}{c}{\textsc{Experiment in Building forest environment}} \\
    \cmidrule(lr){2-5} 
    & \textbf{Inter Collision Rate}&\textbf{\makecell[c]{Environmental \\Collision Rate}}& \textbf{Travel Distance}   &\textbf{Calculation Time(s)}\\
    \midrule
        EDG~\cite{hou2022enhanced}& 0.0521 & 0.083 & 1.115 & 0.042 \\ 
            ECBS~\cite{barer2014suboptimal}& 0.121 & 0.107 & 1.219  & 0.35 \\ 
            \textbf{NTM (Ours)} & \textbf{0.024} & \textbf{0.012} & \textbf{0.994}  & \textbf{0.0025} \\ 
            \midrule
            
            Ground Truth & 0.009  & 0.003 & 0.924 & /  \\ 
            \bottomrule
    \end{tabular}
    }
    \label{tab:results}
\end{table*}

Compared to the baseline methods, the NTM achieves a significant improvement in computation time, being at least 10 times faster (note that the exact number is not meaningful; only the relative magnitude is relevant as the CPU and GPU computation units are different). Moreover, the NTM outperforms the other methods in other evaluation metrics as well, demonstrating the advantages of learning from known trajectory experiences and generalizing beyond them to unseen scenarios in the same environment.

\subsection{Towards Large Scale Trajectories Models}
We have also conducted preliminary experiments to assess the scalability of our approach when dealing with a substantial number of agents. In this particular experiment, we simulated scenarios representing future air traffic.

\begin{figure}[htbp]
  \centering
  \includegraphics[width=0.8\linewidth]{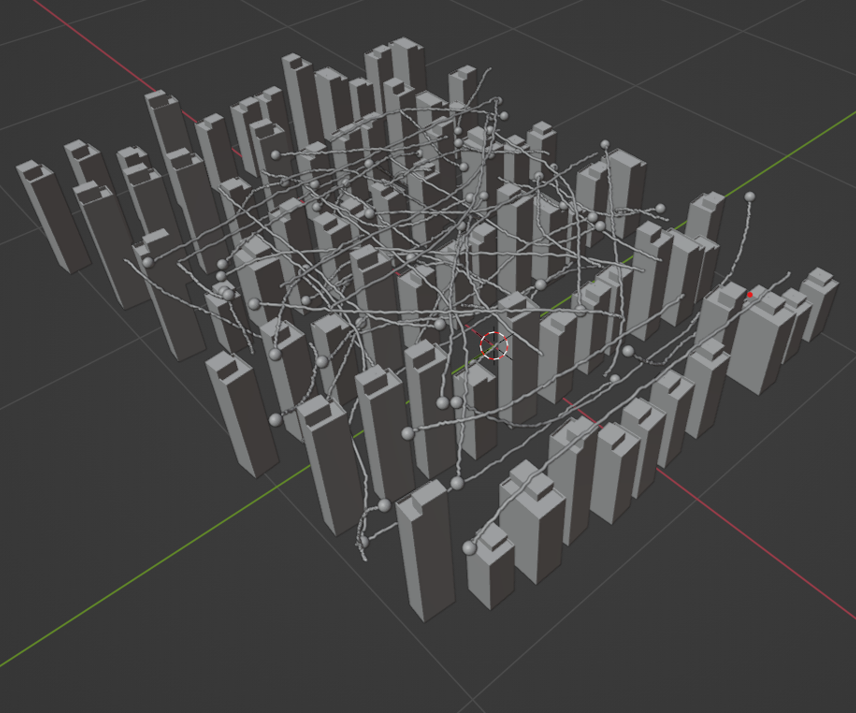}
  \caption{Large-scale trajectory planning in Building Forest Env}
  \label{fig:large}
\end{figure}

Given 64 pairs of start and end positions, the NTM successfully generates collision-free trajectories. We also modify the vanilla transformer~\cite{vaswani2017attention} to support large number coordinates attention across agents\footnote{See our open source code for more details.}. Figure~\ref{fig:large} depicts a screenshot of these scenarios.
\begin{table}[t]
\renewcommand\arraystretch{2}
		\caption{Computation time vs. number of agents}
		\begin{center}
			\begin{tabular}{c| c c c }
				\hline
				& \textbf{Single-agent}& \textbf{Eight-agent}& \textbf{64-agent} \\
				\hline
				\textbf{CT(s)} &0.00253&0.00258&0.00272\\

				\hline
		
			\end{tabular}
			\label{Comparison of time}
		\end{center}
\end{table}

We further compare the computation time as the number of agents increases, as illustrated in Table~\ref{Comparison of time}. As we can see in the table the computation time is not sensitive to the number of agents with our specific implementation of multi-trajectory transformer attention patterns in NTM. This demonstrates the potential for solving large-scale agent trajectory planning problems. We will conduct further research in this direction to fully understand the potential of this approach.

\subsection{Towards Multi-agent Trajectory Coordination Experiments}
\label{sec:coordinate-experiments}
\begin{table}[ht]
\renewcommand\arraystretch{2}
		\begin{center}
		\caption{Multi-Trajectories Coordination}
			\begin{tabular}{c| c c }
				\hline
				\textbf{State} & \textbf{Inter Collision Rate}& \textbf{\makecell[c]{Environmental \\Collision Rate}} \\
				\hline
				\makecell[c]{Before \\Coordination}&0.8688&0.1475\\
				\hline
     \makecell[c]{After \\Coordination} &0.0164 &0.0328 \\
				\hline
		
			\end{tabular}
			\label{Coordination result}
		\end{center}
	\end{table}

Apart from trajectory planning, we conduct experiments in Building Forest to explore the potential usage of the NTM in multi-agent coordination. Given 61 batches trajectories with collisions, we employ the inference process (Algorithm~\ref{alg:neural-traj-inference}) described in Section~\ref{sec:model-inference} with the trajectory proposal step replaced by given trajectory sets. The implementation can solve conflicts and collisions. The coordination result is depicted in Table~\ref{Coordination result}. As we can see in the results, the collision rates are reduced significantly to almost no collisions. More comprehensive experiments will be conducted to fully explore the potential of NTM in this area.

\subsection{Ablation Study on Inference Procedure}
%
\begin{table*}[ht]
    \centering
    \caption{Trajectory performance in Building Forest before and after additional trajectory optimization refinement}
\scalebox{1.1}{
    \begin{tabular}{@{}lcccc@{}}
    \toprule 
    & \multicolumn{4}{c}{\textsc{Experiment in Building Forest environment}} \\
    \cmidrule(lr){2-5} 
    & \textbf{Inter Collision Rate}&\textbf{\makecell[c]{Environmental \\Collision Rate}} &\textbf{ Travel Distance } &\textbf{Calculation Time(s)} \\
    \midrule
        Before Optimization& 0.024 & 0.012 & 0.994 & 0.0025 \\ 
            Five-step Optimization& 0.021 & 0.009 & 0.962 & 0.1652 \\ 
            Ten-step Optimization&0.018  &0.005  & 0.943   &0.3207  \\ 
       
            \bottomrule
   
    \end{tabular}
    }
    \label{tab:ablation study}
\end{table*}
To further illustrate the impact of the additional trajectory optimization refinement, we compare the metrics after performing this refinement with- and without the optimization in the inference process(Algorithm~\ref{alg:neural-traj-inference}). The results are displayed in Table~\ref{tab:ablation study}. From the results, we can observe that the performances of trajectories are improved with further refinement using the trajectory optimizer, demonstrating the effectiveness of trajectory refinement and the possibilities of applying neural trajectories along with an appropriate choice multi-trajectory optimizer to balance between computation speed and solution qualities.

\section{CONCLUSION}

In this study, we propose a neural trajectory model that is capable of generating nearly optimal trajectories for both single-agent and multi-agent scenarios.  We first extend the neural field formulation to accommodate valid and nearly optimal trajectories of complex environments into the implicit neural representation. We further enable the neural trajectories model to support queries of start-goal positions towards multi-trajectories generation. We conduct experiments in simulation and demonstrate that NTM can solve single-agent and multi-agent planning problems. We also conduct experiments demonstrating that NTMs achieve (1) sub-millisecond panning time using GPUs, (2) almost avoiding all environment collisions, (3) almost avoiding all inter-agent collisions, and (4) generating almost shortest paths. We also demonstrate that the same NTM framework can also be used for trajectory correction and multi-trajectory conflict resolution refining low-quality and conflicting multi-agent trajectories into nearly optimal solutions efficiently. To fully utilize the potential of the proposed neural trajectory models, in future work (1) we will conduct a comprehensive study on applying neural trajectory model in large-scale multi-agent planning and coordination problems, (2) we will further extend the neural trajectory framework to accommodate dynamic environment models with sensing inputs, (3) we will also experiment with real-world scenarios adding motion and physical dynamics variables into the trajectories representation to enable physical environment interaction handling.






\bibliography{bib}

\end{document}